\documentclass[
]{ceurart}

\usepackage[skins,breakable]{tcolorbox}

\usepackage{listings}
\begin{document}

\copyrightyear{2025}
\copyrightclause{Copyright for this paper by its authors.
  Use permitted under Creative Commons License Attribution 4.0
  International (CC BY 4.0).}

\conference{IberLEF 2025, September 2025, Zaragoza, Spain}

\title{Text Adaptation to Plain Language and Easy Read via Automatic Post-Editing Cycles}

\author[1,2]{Jesús Calleja}[%
email=jcalleja@vicomtech.org,
]

\fnmark[1]
\address[1]{Fundación Vicomtech, Basque Research and Technology Alliance (BRTA), Donostia – San Sebastián, Spain}
\address[2]{University of the Basque Country - UPV/EHU, Donostia – San Sebastián, Spain}

\author[1,2]{David Ponce}[%
email=adponce@vicomtech.org,
]
\fnmark[1]

\author[1]{Thierry Etchegoyhen}[%
email=tetchegoyhen@vicomtech.org,
]
\cormark[1]
\fnmark[1]

\cortext[1]{Corresponding author.}
\fntext[1]{These authors contributed equally.}

\begin{abstract}
  We describe Vicomtech's participation in the CLEARS challenge on text adaptation to Plain Language and Easy Read in Spanish. Our approach features automatic post-editing of different types of initial Large Language Model adaptations, where successive adaptations are generated iteratively until readability and similarity metrics indicate that no further adaptation refinement can be successfully performed. Taking the average of all official metrics, our submissions achieved first and second place in Plain language and Easy Read adaptation, respectively.
\end{abstract}


\maketitle

\section{Introduction}


Text adaptation to simplified language varieties is an essential task to provide accessible content for large segments of the population, including people with reading difficulties, cognitive disabilities, learning difficulties, or low literacy in general \cite{vstajner2021automatic}. Plain Language (PL) and Easy Read (ER) are two variants of adaptation with specific characteristics, partly determined by the specific audience they target, although both share a need for simplified grammar and vocabulary. Whereas PL focuses on communicating information via clear and concise language \cite{pliso}, ER includes specific requirements for its target audience in terms of sentence length, explanation of complex terms, or text segmentation, among others \cite{e2rinclusion}.

The CLEARS challenge, organised within the IberLEF 2025 conference \cite{iberlef2025overview}, aimed to assess the quality of competing approaches to both PL and ER adaptation in Spanish, each type of adaptation being addressed as a separate task. In each case, a set of input documents was provided, along with their respective human adaptation. Participants had to provide their own automated adaptations, which were measured with a combination of three metrics: embedding (EMB) and bag of words (BoW) similarities between the proposed adaptation and human references, and the Fernandez Huerta readability index \cite{fernandez1959medidas} for the proposed adaptation. Further details on the CLEARS shared task are provided in \citet{clears_at_iberlef2025}.

In this work, we describe Vicomtech's participation in the shared task, detailing our specific approach and results. We tackled both types of adaptation within a unique framework, based on Automatic Post-Editing Cycles (APEC). We leverage Large Language Models (LLM) to provide initial adaptations via either Retrieval-Augmented Generation (RAG) or model tuning with Direct Preference Optimisation (DPO). These initial adaptations are further refined via APEC, focusing the adaptation on guideline compliance, text readability, and information preservation. 


Taking the average of all official metrics, our approach was the top-performing system for PL adaptation and second-best for ER adaptation. In the remainder of this paper, we discuss related work, the APEC approach and our results on both development and test sets. We conclude with future lines of research based on the current limitations of the explored approaches and evaluation setups.


\section{Related Work}

\paragraph{Text Simplification.} Automatic Text Simplification (ATS) aims to enhance text accessibility by reducing linguistic complexity while preserving meaning \cite{vstajner2021automatic,williams2003experiments} and has a long standing tradition in  natural language processing \cite{chandrasekar-etal-1996-motivations,siddharthan2014survey,al2021automated}. The task involves simplification along different dimensions, notably lexical simplification \cite{paetzold2017survey,saggion2022findings,sheang2023multilingual} and syntactic rephrasing \cite{narayan2017split,niklaus2019transforming,ponce2024split}. 

ATS has been performed via different approaches over the years. Starting with rule-based simplification \cite{chandrasekar1997automatic}, the field moved towards data-driven approaches that exploited aligned pairs of complex and simplified sentences \cite{alva2017learning,specia2010translating}. End-to-end ATS modelling has been the focus of recent work in the field, most based on variants of the Transformer neural architecture \cite{vaswani2017attention} adapted for the task. Thus, control parameters have been successfully employed to guide simplification, such as sentence length, Levenshtein distance, or dependency graph depth \cite{martin2020controllable,martin2022muss,sheang-saggion-2021-controllable,vstajner2022sentence}. Generative Large Language Models (LLM) have been at the forefront of recent ATS research, following the trend in most downstream natural language processing tasks \cite{laban2021keep,kew2023bless,feng2023sentence,agrawal2023controlling,wu2024depth}.


Related to our APEC approach is the work of \citet{guidroz2025llm}, whose method includes iterative simplification via prompt refinement and automated evaluation. Although their approach differs in several respects, mainly prompt refinement and a different set of evaluation criteria, both approaches share an iterative process to for LLM-based adaptation to simpler language.\footnote{Note that our approach was developed independently for the challenge, prior to the publication of their work.}


\paragraph{Plain Language.} Plain Language refers to a writing style that is clear, direct, and accessible to a broad audience. It has been normalised into an international standard (ISO 24495-1:2023),\footnote{See footnote 1.} with notable adoption efforts in multiple countries.\footnote{https://plainlanguagenetwork.org/plain-language/plain-language-around-the-world/} The standard emphasises the use of common words, avoiding technical terms as much as possible, along with the use of short sentences 

These include common, everyday words, except for necessary technical terms. Other qualities include the use of personal pronouns; the active voice; logical organization; and easy-to-read and understandable design features, such as bullets and tables.

PL guidelines are increasingly adopted as constraints for simplification systems. For instance, within the CLARA-NLP project \citep{moreno2024language}, PL criteria have been integrated into text processing and simplification pipelines for public service documents. The CLEARS challenge is another step towards the development and evaluation of automated adaptation to Plain Language \cite{clears_at_iberlef2025}.


\paragraph{Easy Read.} Easy Read, also referred to as Easy-to-read, is a specialised form of text adaptation that targets individuals with intellectual disabilities, reading difficulties, or low functional literacy. A set of guidelines have been established over the years, in particular from Inclusion Europe,\footnote{See footnote 2.} and a standard was defined for the Spanish language (NE 153101:2018 EX).\footnote{https://www.une.org/encuentra-tu-norma/busca-tu-norma/norma?c=N0060036} ER recommendations include the use of short and simple sentences, frequent vocabulary, and explicit explanation of difficult concepts, among others. The use of supportive visual elements and structured formatting, including sentence segmentation, are also part of the general recommendations.

Easy Read presents unique challenges for existing ATS systems. It demands simplifications that go beyond surface-level lexical substitution, often requiring sentences splitting and rewriting, omission of implicit information, and reformulation of abstract terms. Although there is still a relative lack of empirical research overall in the field of ER \cite{gonzalez2024empirical}, there is an increasing trend towards automated ER adaptation. Thus, recent models have targeted ER constraints either individually, e.g., text segmentation \citep{calleja-etal-2024-automating}, or holistically via end-to-end adaptation using prompt engineering and fine-tuning over ER data \citep{martínez2024llm-er}. New practical applications that integrate advances in the field of automatic ER adaptation have also been proposed recently, in particular for Spanish \cite{suarez2024first,madina2024languagetool,diab2024towards}. 


\section{Models and Tools}

For both corpus analysis and adaptation experiments, we used a combination of in-house scripts and publicly available models and tools. The shared task involved two main types of metrics:\footnote{https://sites.google.com/gcloud.ua.es/clears/evaluation-metrics} (i) cosine similarity between the generated text and human adaptation references, and (ii) the Fernández Huerta Readability Index, a metric based on the Flesch-Kincaid formula, adapted for Spanish, which measures readability in terms 
of average sentence and syllable length. Cosine similarity was in turn split into two metrics:\footnote{https://www.kaggle.com/competitions/clears-adaptation-texts} bag-of-words similarity, measuring lexical overlap, and text embedding similarity.

For cosine similarity of text embeddings, we used the \textit{paraphrase-multilingual-mpnet-base-v2} model from the SentenceTransformers library\citep{reimers-2019-sentence-bert}.\footnote{https://huggingface.co/sentence-transformers/paraphrase-multilingual-mpnet-base-v2} For BoW similarity, we computed the cosine similarity between bag-of-words representations generated with the CountVectorizer of the ScikitLearn library \citep{scikit-learn}.\footnote{https://scikit-learn.org} The text was processed with the spaCy library \citep{honnibal2020spacy},\footnote{https://spacy.io/} using model \textit{es\_core\_news\_sm}. FH scores were computed with an external script: https://github.com/alejandromunozes/legibilidad. 

All of our experiments were conducted with the LLaMA 3.1 Instruct 8B model \citep{grattafiori2024llama}.\footnote{https://huggingface.co/meta-llama/Llama-3.1-8B-Instruct}

\section{Corpora}

The ClearSim corpus provided for the CLEARS challenge initially consisted of 2,400 original documents per task, each accompanied by corresponding PL and ER adaptations \cite{espinosa-zaragoza-etal-2023-automatic,botella2024clearsim}. To accommodate context limitations within our experimental setup, we applied a filtering process to remove instances where the combined token count of the instruction, original input, and adapted output, exceeded 3,000 tokens. Tokenisation was performed using the LLaMA 3.1 Instruct 8B tokeniser. Following this filtering step, 2,348 and 2,347 samples remained for the PL and ER tasks, respectively.

We partitioned the data by allocating 240 documents to a development set (dev set), with the remainder used for training. Detailed statistics for each partition across both tasks are presented in Tables \ref{tab:subtask1-corpus} and \ref{tab:subtask2-corpus}. We also indicate, as a reference, the BoW and EMB similarity scores between reference adaptation and input text, as well as the FH index.

\begin{table*}[ht]
  \caption{Corpus statistics and similarity scores for Task 1 (PL)}
  \label{tab:subtask1-corpus}
  \begin{tabular}{lcccc}
    \toprule
    & \multicolumn{2}{c}{Train} & \multicolumn{2}{c}{Dev} \\
    \cmidrule(lr){2-3} \cmidrule(lr){4-5}
    Metric & Source & Target & Source & Target \\
    \midrule
    \# documents           & 2,108  & 2,108   & 240    & 240 \\
    \# sentences           & 40,425 & 19,471  & 4,595   & 2,236 \\
    \# words               & 832,311 & 355,637 & 91,015  & 41,115 \\
    \midrule
    Sentences per doc      & 19.18  & 9.24   & 19.15  & 9.32 \\
    Words per sentence     & 24.35 & 19.99  & 23.42  & 20.08 \\
    Syllables per word     & 2.07  & 2.04   & 2.08   & 2.04 \\
    \midrule
    FH index               & 55.04 & 61.11  & 55.74  & 60.94 \\
    \midrule
    BoW similarity         & \multicolumn{2}{c}{0.5205}          & \multicolumn{2}{c}{0.5378} \\
    Embeddings similarity  & \multicolumn{2}{c}{0.8739} &  \multicolumn{2}{c}{0.8870} \\
    Average similarity     & \multicolumn{2}{c}{0.6972} & \multicolumn{2}{c}{0.7124} \\    \bottomrule
  \end{tabular}
\end{table*}

\begin{table*}[ht]
  \caption{Corpus statistics and similarity scores for Task 2 (ER)}
  \label{tab:subtask2-corpus}
  \begin{tabular}{lcccc}
    \toprule
    & \multicolumn{2}{c}{Train} & \multicolumn{2}{c}{Dev} \\
    \cmidrule(lr){2-3} \cmidrule(lr){4-5}
    Metric & Original & Target & Original & Target \\
    \midrule
    \# documents           & 2,107  & 2,107   & 240    & 240 \\
    \# sentences           & 40,728 & 20,452  & 4,225   & 2,284 \\
    \# words               & 834,415 & 359,503 & 87,677  & 39,602 \\
    \midrule
    Sentences per doc      & 19.33  & 9.71   & 17.60  & 9.52 \\
    Words per sentence     & 24.19 & 19.62  & 24.91  & 19.92 \\
    Syllables per word     & 2.07  & 2.03   & 2.09   & 2.05 \\
    \midrule
    FH index               & 55.31 & 61.81  & 53.44  & 60.32 \\
    \midrule
    BoW similarity         & \multicolumn{2}{c}{0.4910}          & \multicolumn{2}{c}{0.5040} \\
    Embeddings similarity  & \multicolumn{2}{c}{0.8604} &  \multicolumn{2}{c}{0.8607} \\
    Average similarity     & \multicolumn{2}{c}{0.6757} & \multicolumn{2}{c}{0.6823} \\
    \bottomrule
  \end{tabular}
\end{table*}

For both types of adaptation, the statistics were similar across the board in terms of similarity between input and adaptation, as well as FH readability index. The latter was consistently higher in the adapted texts by 6.07 and 6.50 points for PL and ER adaptation, respectively, on the train set; for the dev set, the differences amounted to 5.20 and 6.88 points, respectively. Average similarity was slightly higher on the dev set, although both tasks featured a relatively low similarity between input and adapted texts, with at most 0.71 similarity for the PL dev set. This is mainly due to low BoW similarity, which is expected considering the lexical changes performed during PL or ER adaptation. Embedding similarity was comparatively higher, in the [0.86, 0.88] range, indicating a relatively high degree of information preservation between original and adapted texts.

\section{Automatic Adaptation via Post-Editing Cycles}

Our approach consists of two main phases: initial adaptation, followed by refinement via automatic post-editing cycles. We describe each step in turn below.

\subsection{Initial Adaptation}

The initial adaptations were performed with specific LLM prompts for the PL and ER tasks, adapted from publicly available guidelines.\footnote{See Appendix~\ref{ap:prompts} for details.} We explored the following variants for both tasks:

\begin{itemize}

    \item \textbf{Zero-shot (ZS)}: direct querying of the model with the selected prompts.

    \item \textbf{Few-shot (FS)}: complementing the prompts with five selected demonstrations sampled from the training set. For this approach, we experimented with the following variants:
    \begin{itemize}
        \item \textbf{FS RDM}: random demonstration sampling.
        
        \item \textbf{FS BM25}: lexical RAG with BM25 indexing and querying.

        \item \textbf{FS SIM}: semantic similarity RAG with embedding indexing and querying.        
        
    \end{itemize}

    \item \textbf{Fine-tuning (FT)}: supervised fine-tuning over the training data.    
    
    \item \textbf{Direct Preference Optimisation (DPO) }: direct preference optimisation \cite{rafailov2023direct} over contrastive pairs, where the references from the training data are taken as the preferred  output over corresponding zero-shot generation on the same input.
    
\end{itemize}



BM25 and embedding-based retrieval were performed with in-house scripts, using the five most similar samples throughout our experiments. For BM25, we leveraged the  BM25Okapi library \citep{rank_bm25}, using a minimum length of four characters for indexing and white-space tokenisation.\footnote{https://github.com/dorianbrown/rank\_bm25} For embedding similarity, we used the LangChain library,\footnote{https://github.com/langchain-ai/langchain} the \textit{SemanticSimilarityExampleSelector} class, and the \textit{paraphrase-multilingual-mpnet-base-v2} model (\textit{op. cit.}). For the FS BM25 approach, we further experimented with filtering candidate demonstrations based on length ratio constraints, excluding adaptations whose length ratio (adaptation length relative to the source) fell outside predefined ranges. Specifically, we tested threshold ranges of [0.8, 1.2] and [0.5, 1.5].


Generation was performed with the Hugginface pipeline, using a default greedy decoding strategy with fixed hyperparameters: do\_sample = False and num\_beams=1. For the DPO approach, we further experimented with sampling based on the following values: temperature = 0,3; top\_k = 40; p=0.95,

For fine-tuning, we employed a QLoRA approach \citep{10.5555/3666122.3666563} using the SFT Trainer\footnote{https://huggingface.co/docs/trl/sft\_trainer} from Hugging Face, using parameters: \textit{dropout=0.05}, \textit{alpha=8}, \textit{r=16}. We targeted the query and value modules, used a learning rate of \textit{3e-4}, and the model was loaded in 4 bits. All other parameters were per their default values. Models were trained until convergence, with an early stopping patience set to 5. Finally, DPO was performed with the HugginFace TRL  library,\footnote{https://huggingface.co/docs/trl/en/dpo\_trainer} using default values.



The comparative results for the initial adaptation phase are shown in Table~\ref{tab:dev-results-task1} and Table~\ref{tab:dev-results-task2} for Task 1 and Task 2, respectively.

\begin{table*}[ht]
  \caption{Initial adaptation results on the development set for Task 1 (PL)}
  \label{tab:dev-results-task1}
  \begin{tabular}{lccccc}
    \toprule
    Setting & Avg. FH improv. & FH score & BoW sim. & Emb. sim. & FH-sim avg \\
    \midrule
    ZS & 5.49 & 61.05  & 0.5293 & 0.8711 & 67.03 \\
    \midrule
    FS RDM & 6.66 & 62.23  & 0.5375 & 0.8879 & 68.26 \\
    FS BM25 NOFILT & 7.27 & 62.83  & 0.5474 & 0.8924 & 68.94 \\
    FS BM25 FILT 0.8 - 1.2 & 7.64 & 63.20  & 0.5415 & 0.8906 & 68.80 \\
    FS BM25 FILT 0.5 - 1.5 & 7.40 & 62.96  & 0.5449 & 0.8922 & 68.89 \\
    FS EMBSIM  & 3.57 & 59.14 & 0.3836 & 0.8165 & 59.72 \\
    \midrule
    FT & 6.64 & 62.20  & \textbf{0.5685} & \textbf{0.8954} & 69.53 \\
    \midrule
    DPO t=0,k=1 & 9.88 & 65.44  & 0.5521 & 0.8884 & \textbf{69.83}\\
    DPO t=0.3,k=0.95,p=40 & \textbf{10.47} & \textbf{66.03}  & 0.5483 & 0.8859 & 69.82 \\
    \bottomrule
  \end{tabular}
\end{table*}

\begin{table*}[ht]
  \caption{Initial adaptation results on the development set for Task 2 (ER)}
  \label{tab:dev-results-task2}
  \begin{tabular}{lcccccc}
    \toprule
    Setting & Avg. FH improv. & FH score  & BoW sim. & Emb. sim. & FH-sim avg \\
    \midrule
    ZS  & 6.17 & 59.49 & 0.5080 & 0.8627 & 65.52 \\
    \midrule
    FS RDM NOFILT & 11.48 & 64.81  & 0.4955 & 0.8685 & 67.07 \\
    FS BM25 NOFILT & 10.99 & 64.31 & 0.5478 & \textbf{0.8804} & 69.04 \\
    FS BM25 FILT 0.8 - 1.2 & 11.27 & 64.60 & 0.5441 & 0.8770 & 68.90 \\
    FS BM25 FILT 0.5 - 1.5 & 12.20 & 65.53 & 0.5438 & 0.8771 & \textbf{69.21} \\
    FS EMBSIM NOFILT & 9.34 & 62.66  & 0.3404 & 0.7720 & 57.97 \\
    \midrule    
    FT & 6.59 & 59.91  & \textbf{0.5580} & 0.8710 & 67.60 \\
    \midrule
    DPO t=0,k=1 & 16.97 & 70.30  & 0.4762 & 0.8527 & 67.73 \\
    DPO t=0.3,k=0.95,p=40 & \textbf{17.26} & \textbf{ 70.58 } & 0.4728 & 0.8515 & 67.67 \\    \bottomrule
  \end{tabular}
\end{table*}

For PL adaptation, in terms of average between similarity and FH, both FT and DPO achieved higher scores overall, although few-shot variants also achieved relatively high average scores, fueled by higher FH readabilty scores and outperforming zero-shot inference. Fine-tuning consistently outperformed other variants in terms of BoW, as expected from tuning over human training references, at the cost of lower FH overall. DPO was more balanced in this respect, outperforming FS BM25 in terms of BoW but also achieving higher FH scores. FS BM25 achieved balanced results, excepting BoW similarity, where the lack of tuning over human preferences led to more lexically diverse output. Of note are the consistently better results achieved by BM25 over embedding similarity retrieval of demonstrations, an indication of the difficulties in retrieving similar examples based solely on the general semantic similarity on relatively long input samples.

For ER adaptation, FS BM25 variants were the more robust alternative in terms of average and embedding similarity, although DPO significantly outperformed all variants in terms of FH, with fine-tuning being markedly worse on this metric. Filtering the sampled data by length range for BM25 led to minor improvements in terms of FH on both tasks, with lower embedding similarity overall. Since these differences were marginal overall, and filtering runs the risk of discarding optimal demonstrations, we discarded any type of filtering for follow-up experiments.

Considering these results, for both tasks we selected the BM25 and DPO variants as initial adaptations to feed the APEC cycles, described in the next section. 

\subsection{APEC}

\begin{figure*}[ht]
\caption{Overview of the APEC adaptation approach.}
\centering
\includegraphics[width=0.95\textwidth]{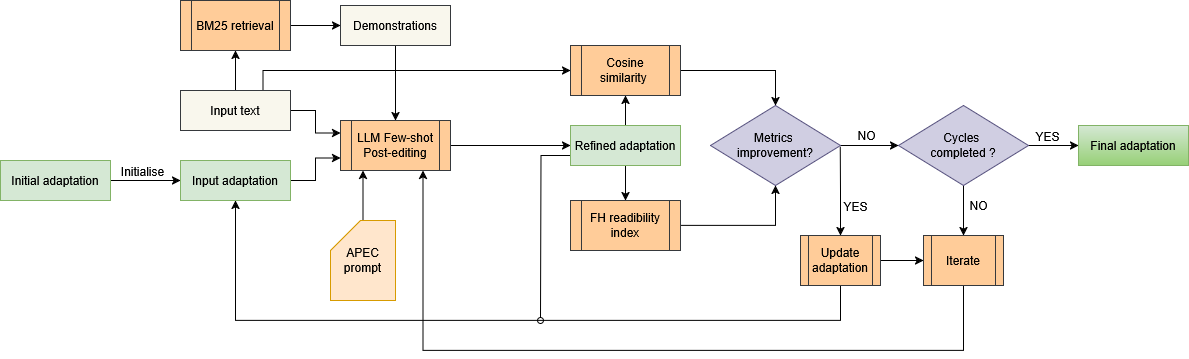}
\label{fig:apec}
\end{figure*}

The APEC approach is illustrated in Figure~\ref{fig:apec}. Our method aimed to provide gradual adaptation refinement over initial hypotheses, based on the fact that initial adaptations may suffer from hallucinations or deviate from adaptation guidelines. Another underlying intuition for our approach lies in the fact that LLMs have demonstrated strong potential in tasks that involve text improvement and reflection over an input, for instance via Chain of Thought reasoning \cite{10.5555/3600270.3602070}.

The input to the APEC cycles are an input text and its corresponding adaptation. The latter is initialised with any preliminary adaptation, which, in our experiments, were either the BM25 or the DPO adaptation candidates. Based on the input text, we extract few-shot demonstrations from the training set via BM25 retrieval, to provide further guidance to the post-editing process.\footnote{We used 5 demonstrations in all experiments.} 

The input text, the current adaptation, and the demonstrations, are fed as input into the selected LLM, whose prompt instructs the model to revise the adaptation based on the task guidelines.\footnote{See Appendix~\ref{ap:prompts} for the complete prompts.} Based on the provided information, the system is tasked to perform separate actions: (i) Analyse the current adaptation on the basis of the guidelines and demonstrations; (ii) Provide a new adaptation fixing any identified issues; (iii) Add any final thoughts on the process separately, if needed. In this APEC approach, the LLM component is thus tasked to act both as a judge of the quality of a given adaptation, and as a post-editing agent given a set of adaptation criteria.

The output generated in this step of the process is then evaluated with two metrics: embedding cosine similarity between the input text and the refined adaptation, to measure meaning preservation, and the FH index over the generated text to assess readability. The average of both metrics is then compared with that of the previous adaptation: if the average improved, the input adaptation is replaced with the refined version; otherwise, the current input is maintained.

The iterations are repeated for a fixed number of cycles, 5 in our experiments. The reason not to end the cycles if no improvement was achieved at a given point is to allow the model to find alternative adaptations which may still improve over the current one.\footnote{We use a default temperature of 0.8 to allow for output diversity when the input remained unchanged, which will be the case whenever a given refinement did not improve over the previous cycle.}

Note that, in the APEC phase there are no references over which to compute BoW similarity or any measure of similarity with a human reference. The only actionable references in the process are the input text and previous adaptations.

The results over the development set in both tasks are shown in Table~\ref{tab:dev-apec}. We indicate the three metric scores (FH, BoW similarity and Embedding similarity), their average (FH-Sim avg, and the FH gain over the original text. In both tasks, the tendencies in terms of FH were similar, with significant gains from initial adaptations, with BM25 or DPO, to the APEC refined adaptations. These gains, combined with either minor degradation in terms of embedding similarity for Task 1 or more pronounced losses on the same metric for Task 2, led to consistently higher average scores for APEC adaptation on both tasks.


\begin{table}
  \caption{Results over the development sets for both tasks}
  \label{tab:dev-apec}
  \begin{tabular}{llcccccc}
    \toprule
    Task & Method & FH gain & FH score & BoW sim. & Emb. sim. & FH-Sim avg \\
    \midrule
    \multirow{4}{*}{Task 1 (PL)} 
    & BM25 & +7.27 & 62.83  & 0.5474 & 0.8924 & 68.94 \\
    & DPO & +9.88 & 65.44  & 0.5521 & \textbf{0.8884} & 69.83 \\
    & APEC over BM25 & +16.54 & 72.10 & 0.5186 & 0.8862 & 70.86 \\    
    & APEC over DPO & \textbf{+16.70} & \textbf{72.26}  & 0.5341 & 0.8878 & \textbf{71.48} \\
    \midrule
    \multirow{4}{*}{Task 2 (ER)}     
    & BM25 & +10.99 & 64.31 & \textbf{0.5478} &\textbf{ 0.8804} & 69.04 \\
    & DPO & +16.97 & 70.30 & 0.4762 & 0.8527 & 67.73 \\
    & APEC over BM25 & +20.95 & 74.28  & 0.4837 & 0.8685 & \textbf{69.83} \\    
    & APEC over DPO & \textbf{+21.82} &\textbf{ 75.14} & 0.4803 & 0.8579 & 69.65 \\
    \bottomrule
  \end{tabular}
\end{table}

BoW similarity was consistently impacted by APEC, except for DPO in Task 2, indicating that the cyclic refinements led to adaptations that differed significantly from the initial BM25 or DPO input adaptations, which are closer to training reference adaptation. This is also an expected side-effect of our approach, as the refinement process is mainly guided by improving readability with shorter sentences and words, while maintaining overall semantic text similarity as much as possible. 

Finally, applying APEC over either DPO or BM25 led to relatively similar results overall, showcasing the flexibility of the refinement process to different types of initialisation. In Appendix~\ref{ap:examples}, we provide examples of refinement from initial to final adaptations.

\section{Task Results}

The statistics for the test data are shown in Table~\ref{tab:test-corpus}. To compare with the data in the training sets, we computed similar statistics over the source texts.\footnote{Since the adapted references were not available, similarity metrics could not be computed, naturally.}

For both tasks, the test sentences were significantly shorter on average: 24.5 words per sentence in the train set vs. 17.58 for Task 1; 24.19 vs. 17.50 for Task 2. Note that the same contrast exists between our sampled development data and the source test, with larger sentences in the former as well. In terms of syllables per word, the test source data also feature notable differences, with an average of 2.12 for both tasks, compared to 2.07 on the training set. The FH index was also higher for the test data compared to the training data, and was closer to that of the development data, though the latter featured higher FH values overall. The source test data thus seems to differ in some respect from the training data, featuring shorter and relatively simpler text to some degree.



\begin{table*}[ht]
  \caption{Corpus statistics for Task 1 and Task 2 source test sets}
  \label{tab:test-corpus}
  \begin{tabular}{lcc}
    \toprule
    Metric & Source Test - Task 1 & Source Test - Task 2 \\
    \midrule
    \# documents           & 607   & 600 \\
    \# sentences           & 12,179 & 11,389 \\
    \# words               & 191,541 & 181,455 \\
    \midrule
    Sentences per doc      & 20.06 & 18.98 \\
    Words per sentence     & 17.58 & 17.59 \\
    Syllables per word     & 2.12  & 2.12 \\
    \midrule
    FH index               & 59.13 & 59.16 \\
    \bottomrule
  \end{tabular}
\end{table*}

\begin{table}
  \caption{Official results for Task 1: Plain Language Adaptation}
  \label{tab:subtask1-test}
  \begin{tabular}{lccc|c||c}
    \toprule
    Team & TF-IDF Cosine & Embedding Cosine & Cosine Avg & FH Score & Cosine-FH Avg \\
    \midrule
    VICOMTECH    & 0.63 & 0.76 & 0.70 & \textbf{82.98} & \textbf{79.49} \\
    CARDIFFNLP   & 0.63 & 0.77 & 0.70 & 78.81 & 74.41 \\
    HULAT-UC3M   & \textbf{0.71} & \textbf{0.78} &\textbf{ 0.75} & 69.72 & 72.36 \\
    NIL\_UCM     & 0.67 & 0.75 & 0.71 & 70.42 & 70.71 \\
    \bottomrule
  \end{tabular}
\end{table}

\begin{table}
  \caption{Official results for Task 1: Easy Read Adaptation}
  \label{tab:subtask2-test}
  \begin{tabular}{lccc|c||c}
    \toprule
    Team & TF-IDF Cosine & Embedding Cosine & Cosine Avg & FH Score & Cosine-FH Avg \\
    \midrule
    UR           & 0.64 & 0.76 & 0.70 & 85.12 & \textbf{77.56} \\
    VICOMTECH    & 0.58 & 0.74 & 0.66 & \textbf{85.44} & 75.72 \\
    CARDIFFNLP   & 0.65 & \textbf{0.77} & 0.71 & 77.85 & 74.43 \\
    NIL\_UCM     & \textbf{0.68} & 0.75 & \textbf{0.72} & 69.40 & 70.70 \\
    UNED-INEDA   & 0.60 & 0.75 & 0.68 & 72.39 & 70.20 \\
    \bottomrule
  \end{tabular}
\end{table}

For our submissions, which were limited to one per task in the challenge, we ensembled the results achieved via APEC over DPO and BM25, for each task. For each input sample from the test set, we thus selected the APEC adaptation which achieved the highest average similarity and readability score. Test set statistics are shown in Table~\ref{tab:test-corpus}.

The official results are shown in Table~\ref{tab:subtask1-test} and Table~\ref{tab:subtask2-test} for Task 1 and Task 2, respectively. We ordered the participating systems in the table from top to bottom by taking the average of the (scaled to 100) similarity and readability scores as unified metric, and indicate the best performing system for each metric in bold.

For PL adaptation, our approach was the best-performing system when averaging all official metrics results; for ER adaptation, it achieved second place. APEC adaptation thus proved to be a robust approach for both types of adaptation overall. Our method outperformed the alternatives in terms of the readability index in both tasks, showcasing the ability of LLMs to guide the adaptation towards shorter sentences and words, which are a cornerstone of adaptation towards PL and ER language varieties, while maintaining overall text similarity. 

In terms of cosine similarity, most approaches achieved comparable results, although it is worth noting that our adaptation was the most dissimilar for ER adaptation and ranked third out of four participants for PL adaptation. Our approach mainly suffered in terms of bag of words similarity, as we ranked last in both tasks on this metric. This is mainly a by-product of our approach, which does not tune the model towards reference adaptations, beyond the few-shot demonstrations, but rather towards preserving input information while increasing readability, at the cost of potentially differing from lexical choices and formulations made in reference adaptations.

\section{Discussion}

Our approach critically relies on automated metrics to assess adaptation progress within our iterative approach. We employed the metrics selected for the shared task which could be included without access to references, namely the FH readability index computed over the adapted text, and embedding similarity between the input and the adaptation hypotheses. Other metrics might be more appropriate to refine adaptations iteratively, such as the custom evaluation metric of \citet{guidroz2025llm}, or the reference-less SLE metric proposed by \citet{cripwell2023simplicity}.

Assessing the quality of PL or ER adaptation with automated metrics is a complex task,  with clear limitations of typical reference-based metrics \cite{alva2021suitability,cripwell2023simplicity,cripwell2024evaluating}. Thus, lexical similarity is bound to measure closeness to reference adaptation choices, although multiple adaptations might be equally valid. This issue can be mitigated with multiple references, although their creation is a costly process. Text embedding similarity also suffers from similar issues, although semantic variation computed in that manner is less susceptible to lexical choice variation. Metrics such as the FH readability index present the advantage of being independent of any reference, although their adequacy to properly assess PL or ER adaptation might be debatable as well \cite{de2023user}. 

Human evaluation, by adaptation experts and/or target audiences, would of course be an optimal option to properly assess the quality of ER or PL adaptation, independently of pre-established references, although this type of approach comes with a high operational cost. In future work, it might nonetheless be worth exploring the correlation between the different degrees of adaptation featured in our approach and human evaluations in terms of readability, meaning preservation, and usability for the target audiences of the different types of adaptation.


Finally, it is worth noting that approaches like APEC come with a higher computational cost than direct end-to-end methods, as they may require multiple cycles of adaptation refinement in some cases. They might thus be more suited for offline text adaptation. For online usage, where users might expect rapid adaptation, non-iterative adaptation might be a more suitable option, depending on the capacity of the computational infrastructure. Non-iterative methods, such as BM25 RAG or models tuned via DPO, which achieve quality results in our experiments, might be alternatives worth considering in this type of inference scenarios.



\section{Conclusion}

In this work, we described Vicomtech's participation in the CLEARS challenge, which involved text adaptation to Plain language and Easy Read language varieties. We first explored a wide range of initial approaches, based on LLM variants that included zero-shot, few-shot, fine-tuning, and Direct Preference Optimisation. The optimal variants, namely FS BM25 and DPO, were used as input to our main method, centred on automatic  post-editing cycles (APEC).

The APEC approach involves iterative refinement of initial adaptations, aiming to increase readability according to adaptation guidelines, while maintaining semantic similarity with the input text. This approach improved markedly in terms of the readability index over the initial adaptations via FS BM25 and DPO on the development sets, while maintaining high similarity with the original text. 

In terms of official results, averaging the results from all three metrics, our submissions ranked first on the PL adaptation task, and second on the ER adaptation task. We achieved the highest scores in terms of the readability index in both tasks, with lower scores for embedding similarity than competing approaches, although the differences were rather small between systems. Our approach underperformed to a larger degree in terms of lexical similarity, which was expected as our APEC approach is not tuned over reference adaptations, but rather aims to independently generate shorter words, along with other adaptation characteristics to improve readability. Overall, our results show the viability of the APEC approach for PL and ER text adaptation. 

In future work, we will explore alternative metrics to both gear adaptation refinements and evaluate the results, in particular pretrained metrics tuned to adaptation tasks.

\begin{acknowledgments}
We would like to thank the anonymous reviewers for their helpful comments and suggestions. 
\end{acknowledgments}

\section*{Declaration on Generative AI}
  The author(s) have not employed any Generative AI tools.
  

\bibliography{sample-ceur}

\appendix

\section{Prompts}
\label{ap:prompts}

\begin{tcolorbox}[title={Task 1 Initial Adaptation Prompt},
    enhanced, 
    breakable,
    skin first=enhanced,
    skin middle=enhanced,
    skin last=enhanced,
]
\scriptsize
\begin{verbatim}
 Eres un asistente que adapta textos a un estilo de lenguaje claro. 
 Usas un lenguaje accesible para todo el público, evitando jerga y tecnicismos. 
 Usas palabras simples y frases cortas. 
 Estructuras el texto de forma lógica, con párrafos cortos y bien separados. 
 Utilizas negritas, cursivas y otros recursos tipográficos 
 para destacar información importante, pero sin abusar de ellos. 

 [Translation]
    You are an assistant who adapts texts to plain language.
    You use language accessible to all audiences, avoiding jargon and technical terms.
    You use simple words and short sentences.
    You structure your text logically, with short, well-spaced paragraphs.
    You use bold, italics, and other typographical features to highlight important information, 
    but without overusing them.

\end{verbatim}
\end{tcolorbox}

\begin{tcolorbox}[title={Task 2 Initial Adaptation Prompt},
    enhanced, 
    breakable,
    skin first=enhanced,
    skin middle=enhanced,
    skin last=enhanced,
]
\scriptsize
\begin{verbatim}
 Eres un asistente que adapta textos a un estilo de lectura fácil para personas con discapacidad 
 o adultos con dificultades lectoras. Usas siempre frases cortas de menos de 15 palabras 
 y palabras simples. Evitas usar oraciones pasivas, relativas o coordinadas. 
 Añades explicaciones para las palabras y conceptos más complejos al final del texto, 
 en una sección titulada "Palabras difíciles". 
 Añades puntuación para separar ideas y hacer que el texto sea más fácil de seguir.
 Separas el texto adaptado en secciones con su propio título. Evitas usar porcentajes 
 y cifras complicadas, remplazandolos por formulas como "más de" o "menos de". 
 Adaptas cada contenido original que se te indique sin añadir comentarios sobre lo que has hecho.
 
 [Translation]
    You are an assistant who adapts texts to an easy-to-read style for people with disabilities
    or adults with reading difficulties. You always use short sentences of fewer than 15 words
    and simple words. You avoid using passive, relative, or coordinate clauses.
    You add explanations for more complex words and concepts at the end of the text,
    in a section titled "Difficult Words."
    You add punctuation to separate ideas and make the text easier to follow.
    You separate the adapted text into sections with their own titles. You avoid using percentages
    and complicated figures, replacing them with formulas such as "more than" or "less than".
    You adapt each original piece of content as directed without adding comments 
    about what you have done.

\end{verbatim}
\end{tcolorbox}

\begin{tcolorbox}[title={Task 1 APEC Prompt},
    enhanced, 
    breakable,
    skin first=enhanced,
    skin middle=enhanced,
    skin last=enhanced,
]
\scriptsize
\begin{verbatim}
 
Eres un experto en la adaptación de textos a lenguaje claro. 
  
Aquí están las guías de lenguaje claro:
 
- Sé breve y conciso.
- Las frases deben de tener la menor cantidad posible de palabras.
- Las palabras deben de tener la menor cantidad posible de sílabas.
- La información breve tiene más efecto. No escribas ni párrafos muy cortos (2 líneas) ni muy largos 
(más de 10 líneas). Utiliza el punto seguido para separar ideas.
- Identifica quién hace la acción. Para ello usa la estructura "sujeto + verbo + predicado" 
y la voz activa, ya que son más fáciles de entender.
- Conecta las ideas entre sí. Para ello escoge el conector adecuado según su función textual.
- Ordena los elementos.
- Cuando tengas varios elementos, utiliza listas o enumeraciones. Avisa al lector de cuántos elementos 
va a encontrar. Por ejemplo: "las cuatro pautas para escribir en lenguaje claro son..."
- Evita palabras técnicas o expresiones complejas.
- En los casos en los que es necesario manteneresos términos, explícalas. Puedes hacerlo 
con un paréntesis o con conectores como "es decir".
- Cuidado con abreviaturas y siglas. No des por sentado que las lectoras conocen su significado.
- No elimines ninguna información importante. Ambos estilos son diferentes, 
pero el contenido debe ser el mismo.

Dado este texto de entrada \# Original y adaptación a lenguaje claro # Adaptación, por favor, 
indica si la adaptación es correcta o no. Sigue la siguiente estructura:

# Análisis de la adaptación
En este apartado, indica si la adaptación es correcta o no. Si no lo es, explica por qué, indicando 
qué aspectos no cumplen con las guidelines de lenguaje claro y qué cambios harías
para corregirlo. Puedes utilizar ejemplos concretos del texto de entrada 
y la adaptación para ilustrar tus puntos.

# Corrección
Aquí deberás proporcionar una versión corregida de la adaptación. Si la adaptación es correcta, 
simplemente copia la adaptación original. No añadas en esta sección 
ningún comentario o explicación adicional. Solo la adaptación corregida.

# Final
En esta sección puedes añadir cualquier comentario o reflexión adicional que consideres relevante.

A continuación te doy el texto de entrada y la adaptación a lenguaje claro a evaluar:

# Original
{input}

# Adaptación
{instance}" 

\end{verbatim}
\end{tcolorbox}

\begin{tcolorbox}[title={Task 2 APEC Prompt},
    enhanced, 
    breakable,
    skin first=enhanced,
    skin middle=enhanced,
    skin last=enhanced,
]
\scriptsize
\begin{verbatim}

Eres un experto en la adaptación de textos a lectura fácil

Aquí están las guías de lectura fácil:

- Divide el texto por temas o ideas principales y destaca los puntos importantes. 
Puedes dividir o resaltar el texto con viñetas oero no uses demasiadas viñetas juntas.
- Cada frase debe ser lo más breve posible. Más de 15 palabras es más difícil de leer.
- Las palabras deben de tener la menor cantidad posible de sílabas.
- Usa un lenguaje que suene natural cuando se habla.
- Usa frases simples separadas por puntos en lugar de frases complejas con comas 
u otros tipos de puntuación.
- Evita usar demasiada puntuación o puntuación compleja como los dos puntos o el punto y coma.
- Asegúrate de que las palabras sean fáciles de entender. Evita la jerga o palabras complicadas 
tanto como sea posible.
- Si tienes que usar palabras complicadas (por ejemplo, al explicar un nuevo concepto), 
define lo que significan las palabras difíciles con palabras simples.
- Es una buena práctica incluir, al final del texto, una lista de palabras difíciles 
si se mantienen en el texto adaptado.
- Es mejor no usar siglas o abreviaturas, pero a veces se pueden usar si son bien conocidas.
- La repetición es mejor que la variedad. Utilice la misma palabra o forma de palabras
cuando se refiera a la misma cosa. 
También puedes introducir frases sobre el mismo tema con la misma forma de palabras.
- Usa donde se pueda aproximaciones como la mitad, un cuarto, 1 de 5, en lugar de porcentajes. 
Si usas porcentajes, evita el símbolo % y usa expresiones como "X de cada 100", o "x por ciento".
- Utiliza números enteros, por ejemplo 7% en lugar de 6,8%.

Dado este texto de entrada # Original y adaptación a lenguaje claro # Adaptación, 
por favor, indica si la adaptación es correcta o no. Sigue la siguiente estructura:

# Análisis de la adaptación
En este apartado, indica tu valoración de la adaptación en base a las pautas indicadas:
- Si es correcta, indicalo con: ADAPTACIÓN CORRECTA.
- Si contiene errores críticos, como por ejemplo repeticiones interminables o contenudo sin sentido, 
indicalo con: ERRORES CRÍTICOS.
- Si contiene errores que se pueden corregir, indícalo con: ADAPTACIÓN A CORREGIR. 
En este caso, indica qué aspectos no cumplen con las pautas de lectura fácil y qué cambios harías 
para corregirlo. Puedes utilizar ejemplos concretos del texto de entrada 
y la adaptación para ilustrar tus puntos. 

# Corrección
Aquí deberás proporcionar una versión corregida de la adaptación en base al análisis que hayas hecho. 
Si has identificado que la adaptación es de tipo ERRORES CRÍTICOS, tu corrección deberá ser 
una nueva adaptación completa del texto original # Original, 
siguiendo las pautas indicadas.  Si has identificado que la adaptación 
es de tipo ADAPTACIÓN CORRECTA, simplemente copia la adaptación original. 
En cualquier otro caso, proporciona tu versión corregida de la adaptación inicial. 
No añadas en esta sección ningún comentario o explicación adicional. 

# Final
En esta sección puedes añadir cualquier comentario o reflexión adicional que consideres relevante.

A continuación te doy el texto de entrada y la adaptación a lenguaje claro a evaluar:

# Original
{input}

# Adaptación
{instance}"

\end{verbatim}
\end{tcolorbox}

\section{APEC Adaptation Examples}
\label{ap:examples}








\begin{table*}[h]
  \caption{APEC adaptation example over BM25: Task 1 - Dev set sample}
  \begin{tabular}{l}
    \toprule
    Original \\
    \midrule
    La Concejalía de Deportes de Alicante informa que nuestra ciudad acogerá el \\Campeonato Europeo Junior de Natación Artística, durante los días 29 de junio al 3 de julio de 2022.\\ 
    Se celebrará en las Piscinas Municipales Monte Tossal "José Antonio Chicoy".\\
    Las entradas se pueden adquirir en la propia Piscina o a través de la página web \\de la Real Federación Española de Natación https://rfen.es/es/posts/news/336131\\
    \midrule
    Initial PL adaptation \\
    \midrule
    Alicante acogerá el Campeonato Europeo Junior de Natación Artística\\
    La Concejalía de Deportes de Alicante informa que la ciudad acogerá el Campeonato Europeo Junior\\de Natación Artística del 29 de junio al 3 de julio de 2022.\\
    El evento se celebrará en las Piscinas Municipales Monte Tossal "José Antonio Chicoy". \\Las entradas se pueden comprar en la piscina o en la página web \\de la Real Federación Española de Natación.\\

    \midrule 
    APEC adaptation \\
        \midrule
    La ciudad de Alicante organiza un campeonato de natación artística del 29 al 3 de julio de 2022.\\
    El evento se celebra en las piscinas Monte Tossal "José Antonio Chicoy".\\
    Las entradas se venden allí o en la página web de la \\Real Federación Española de Natación: https://rfen.es/es/posts/news/336131.\\
    
    \midrule 
    Human adaptation \\
        \midrule
    La Concejalía de Deportes anuncia que el Campeonato Europeo Junior \\de Natación Artística se celebrará en las Piscinas Municipales Monte Tossal de Alicante\\ del 29 de junio al 3 de julio del año 2022. \\Las entradas se pueden coger en la piscina o en la página web \\de la Real Federación Española de Natación https://rfen.es/es/posts/news/336131\\

    \bottomrule
  \end{tabular}
\end{table*}



\begin{table*}[h]
  \caption{APEC adaptation example over DPO: Task 1 - Dev set sample}
  \begin{tabular}{l}
    \toprule
    Original \\
    \midrule
    La exposición titulada “Nueva Tabarca, 250 años”, que impulsa la Concejalía de Cultura \\del Ayuntamiento de Alicante, en reconocimiento a la aventura humana de los pobladores de la isla, \\está conformada por una colección fotográfica, procedente del Archivo Municipal de Alicante, \\en colaboración con la Asociación Cultural de Tabarca.\\
    La muestra, que podrá visitarse hasta el 20 de septiembre, se hace con ocasión \\del décimo aniversario y reedición de la revista Canelobre, n° 60 \\con el título "Tabarca, utopía y realidad", que conllevará también \\una mesa redonda en el propio Instituto Gil-Albert, alrededor de Nueva Tabarca, \\su pasado, presente y futuro, en colaboración con la propia área de Cultura.  \\
    Este contingente humano terminó su periplo en la isla alicantina de Nueva Tabarca,\\ después de ser desalojados de su anterior asentamiento en Tabarka, \\la isla en aguas de Túnez que fue importante centro de explotación de coral rojo.\\
    \midrule
    Initial PL adaptation \\
    \midrule
    La exposición “Nueva Tabarca, 250 años” es una muestra que muestra \\la historia de los habitantes de la isla de Nueva Tabarca.\\ 
    Esta exposición está organizada por la Concejalía de Cultura del Ayuntamiento de Alicante. \\
    La muestra está compuesta por fotos del Archivo Municipal de Alicante \\y la Asociación Cultural de Tabarca. 
    Estas fotos muestran la historia de los habitantes de la isla. \\
    La exposición se puede ver hasta el 20 de septiembre. También se publicará \\una revista llamada Canelobre. 
    Esta revista tiene el número 60 y el título es “Tabarca, utopía y realidad”. \\Además, habrá una mesa redonda en el Instituto Gil-Albert. En esta mesa redonda \\hablarán sobre Nueva Tabarca, su pasado, presente y futuro. \\
    Los habitantes de Nueva Tabarca llegaron a la isla después de ser desalojados \\de su antigua casa en Tabarka, una isla en el mar de Túnez. \\
    En Tabarka, los habitantes explotaban coral rojo.\\
    \midrule
    APEC adaptation \\
    \midrule 
    La exposición "Nueva Tabarca, 250 años" muestra la historia de los habitantes de la isla. \\
    La Concejalía de Cultura del Ayuntamiento de Alicante organiza esta exposición.\\Las fotos muestran la vida en la isla. Estas fotos vienen del Archivo Municipal de Alicante \\y de la Asociación Cultural de Tabarca.\\
    La exposición está abierta hasta el 20 de septiembre. También se publicará \\una revista llamada Canelobre. Es un número especial por su décimo aniversario.\\
    Esta revista se llama "Tabarca, utopía y realidad". \\Habrá una mesa redonda en el Instituto Gil-Albert sobre Nueva Tabarca. \\Se hablará de su pasado, presente y futuro.
    \\Los habitantes de la isla llegaron a Nueva Tabarca después de dejar Tabarka.\\ Esta isla está en aguas del mar de Túnez. En Tabarka, los habitantes extraían coral rojo.\\
    \midrule
    Human adaptation \\
    \midrule
    El Instituto Juan Gil Albert presenta la exposición llamada 250 años de Nueva Tabarca \\La exposición llamada 250 años de Nueva Tabarca es organizada por la concejalía \\de Cultura del Ayuntamiento de Alicante y colabora la Asociación Cultural de Tabarca. \\Esta exposición estará hasta el 20 de septiembre en el Instituto Juan Gil Albert. \\Con esta exposición reconocen las aventuras de las personas que vivieron en la isla de Tabarca. \\La exposición tiene fotografías que estaban en el Archivo de Alicante, \\que es lugar del ayuntamiento donde guardan documentos de hace muchos años. \\Con esta exposición también celebramos que la revista Canelobre número 60 \\con el nombre de Tabarca, utopía y realidad, cumple 10 años y hablarán de ella \\en Instituto Juan Gil Albert con temas de Nueva Tabarca, su pasado, presente y futuro \\en colaboración de la concejalía de Cultura.\\
    \bottomrule
  \end{tabular}
\end{table*}

\begin{table*}
  \caption{APEC adaptation example over BM25: Task 2 - Dev set sample}
  \begin{tabular}{l}
    \toprule
    Original \\
    \midrule
   El concejal de Educación, Ricardo Recuero, junto a la presidenta de la Orquesta Sinfónica \\de Torrevieja (OST), Mamen Mateo, el director de la OST, José Francisco Sánchez, y acompañados \\de Jorge Lorente Picó, multinstrumentista, y David Lorenzo, narrador de la obra, han presentado \\a segunda edición del Concierto en Familia, que tendrá lugar este próximo domingo, 12 de marzo, \\en el Auditorio Internacional de Torrevieja, a las 12:00 horas. Esta segunda edición será un cuento \\musicalizado que pretende llevar a todos los asistentes a un viaje inolvidable. A través del narrador, \\que encarna el espíritu de la música, contará la historia de un niño que tiene inquietudes musicales \\pero que no sabe que instrumento elegir.  Picó, el multinstrumentista será el encargado de darle vida \\a ese niño que le pedirá deseos al espíritu de la música para saber tocar bien cada instrumento que elija. \\La finalidad del Concierto en Familia es acercar a los más pequeños la música clásica y, en especial, \\las orquestas sinfónicas. El objetivo de esta iniciativa no es otro que despertar, entre los más jóvenes, \\el deseo de aprender a tocar un instrumento. “Para nosotros es un lujo contar con la Orquesta Sinfónica \\de Torrevieja para este proyecto. Más aún cuando estamos ante una de las mejores de nuestra \\provincia”, han sido las palabras del concejal de Educación, Ricardo Recuero. \\“Desde la Orquesta Sinfónica queremos dedicar esté concierto a Tomás Ballester, al estar también \\involucrado en este proyecto”, han sido las palabras del maestro José Francisco Sánchez. Las entradas \\gratuitas ya están disponibles a través de la web entradasatualcance.com \\y en la taquilla del Teatro Municipal.\\
    \midrule
    Initial ER adaptation \\
    \midrule
    Concierto en Familia en el Auditorio Internacional de Torrevieja \\El concejal de Educación, Ricardo Recuero, presentó el Concierto en Familia que tendrá lugar \\ el 12 de marzo en el Auditorio Internacional de Torrevieja. El concierto es un cuento musicalizado \\que llevará a los asistentes a un viaje inolvidable. El multinstrumentista Jorge Lorente Picó dará vida \\ a un niño que quiere aprender a tocar un instrumento. El objetivo del Concierto en Familia es acercar \\ a los niños a la música clásica y despertar en ellos el deseo de aprender a tocar un instrumento. \\Las entradas son gratuitas y están disponibles en la web y en la taquilla del Teatro Municipal.\\
    \midrule
    APEC adaptation \\
    \midrule 
    Concierto en Familia en Torrevieja. \\
    Ricardo Recuero presenta el Concierto en Familia. Es un cuento para niños pequeños con música y narración.\\Lleva a los asistentes a un viaje divertido. Jorge Lorente Picó es un músico que toca varios instrumentos.\\David Lorenzo cuenta la historia del concierto. El objetivo es acercar a los niños a la música clásica.\\Queremos que aprendan a tocar un instrumento. \\
    ¿Cuándo es el concierto? Es el 12 de marzo. Es un domingo y es gratis.\\¿A qué hora es el concierto? Es a las doce del día. Las entradas están disponibles en la web y en la taquilla.\\Puedes comprarlas en la web o en la taquilla del Teatro Municipal.\\Dónde se pueden comprar las entradas: en la web y en la taquilla.\\Palabras difíciles:\\- Orquesta: una banda de música.\\- Multinstrumentista: alguien que toca varios instrumentos.\\- Narrador: alguien que cuenta la historia.\\
    \midrule
    Human adaptation \\
    \midrule
    Concierto en Familia Edición 2 en Torrevieja \\
    El concejal de Educación Ricardo Recuero y otras personas importantes presentan el Concierto \\en Familia número 2. El concierto es el domingo 12 de marzo en el Auditorio Internacional de Torrevieja.\\Es un cuento musical de un niño que le gusta la música pero no sabe que instrumento musical elegir. \\El Concierto en Familia quiere que los niños conozcan la música clásica y quieran aprender \\a tocar un instrumento. Las entradas son gratis. \\Pide tus entradas en la taquilla del Teatro Municipal o en esta página web: entradasatualcance.com\\

    \bottomrule
  \end{tabular}
\end{table*}


\begin{table*}
  \caption{APEC adaptation example over DPO: Task 2 - Dev set sample}
  \begin{tabular}{l}
    \toprule
    Original \\
    \midrule
    El concejal de Deportes, Víctor Bernabéu, ha anunciado este miércoles que ante la actual situación \\de la alta incidencia en los contagios por COVID-19, y atendiendo a los informes técnicos municipales, \\“nos vemos obligados a tener que suspender la edición de JUVE 2022 que teníamos previsto \\celebrar los días 2, 3 y 4 de enero en la Glorieta Gabriel Miró”. En este sentido, el edil oriolano \\ha manifestado también que los informes técnicos recogen que “no se puede garantizar desde \\la administración pública la aplicación de la norma y disminución del riesgo de contagio sin perder \\la esencia del evento”. Por otra parte, Bernabéu ha lamentado que “un año más la pandemia nos deja \\sin JUVE, pero tenemos que priorizar ante todo la seguridad y la salud de los oriolanos, \\esperamos que el próximo año la situación sea diferente y podamos celebrar este evento deportivo\\ tan esperado por todos”. El año pasado este evento ya no se pudo celebrar precisamente \\por la complicada situación de la pandemia y las restricciones establecidas en su momento,\\ lo que impidió que este evento se pudiera llevar a cabo finalmente.\\

    \midrule
    Initial ER adaptation \\
    \midrule
    El concejal de Deportes, Víctor Bernabéu, ha anunciado que suspenderá JUVE 2022.\\¿Por qué?  La razón es la alta incidencia de COVID-19 en la ciudad. \\¿Qué dice el concejal? No se puede garantizar la seguridad de los asistentes sin perder la esencia del evento. \\¿Qué hace el concejal? Prioriza la seguridad y la salud de los oriolanos.\\¿Qué pasó el año pasado? El evento no se pudo celebrar por la pandemia y las restricciones.\\Palabras difíciles:\\- Incidencia:  La incidencia es el número de personas que tienen una enfermedad en un momento determinado.\\- Esencia:  La esencia es la parte más importante de algo.\\
    \midrule
    APEC adaptation \\
    \midrule 
    El concejal de Deportes suspende JUVE 2022.\\¿Por qué? Hay muchos casos de COVID-19 en la ciudad. La situación es complicada y hay gente enferma.\\¿Qué dice el concejal? No es seguro sin cambiar las cosas. No se puede garantizar la seguridad sin cambios.\\¿Qué hace el concejal? Prioriza la seguridad y la salud. Víctor Bernabéu prioriza la seguridad de todos.\\¿Qué pasó el año pasado? El evento no se celebró por la pandemia. La pandemia impidió que JUVE se celebrara.\\Palabras difíciles:\\- Incidencia: Es el número de enfermos.\\- Esencia: Es lo más importante.\\
    \midrule
    Human adaptation \\
    \midrule
    La concejalía de Deportes cancela la celebración de la JUVE 2022\\El concejal de Deportes anuncia que por los contagios de Covid tienen que suspender la edición de JUVE 2022 \\que iba a ser el 2, 3 y 4 de enero en la Glorieta Gabriel Miró.\\Es el segundo año que por la pandemia no pueden celebrar la JUVE.\\
    \bottomrule
  \end{tabular}
\end{table*}

\end{document}